\documentclass[10pt,twocolumn,letterpaper]{article}

\usepackage{cvpr}              % To produce the CAMERA-READY version
\usepackage{multirow}

\definecolor{cvprblue}{rgb}{0.21,0.49,0.74}
\usepackage[pagebackref,breaklinks,colorlinks,allcolors=cvprblue]{hyperref}

\def\paperID{13233} % *** Enter the Paper ID here
\def\confName{CVPR}
\def\confYear{2025}

\title{High Temporal Consistency through Semantic Similarity Propagation in Semi-Supervised Video Semantic Segmentation for Autonomous Flight}

\author{Cédric Vincent$^{1,2,*}$    Taehyoung Kim$^{2,*}$    Henri Meeß$^{2}$\\
$^{1}$Télécom Paris, Institut Polytechnique de Paris     $^{2}$Fraunhofer IVI\\
{\tt\small 13vincentcedric@gmail.com,    \{taehyoung.kim, henri.meess\}@ivi.fraunhofer.de}
}

\begin{document}
\maketitle
\def\thefootnote{*}\footnotetext{Equal contribution}
\begin{abstract}
Semantic segmentation from RGB cameras is essential to the perception of autonomous flying vehicles. The stability of predictions through the captured videos is paramount to their reliability and, by extension, to the trustworthiness of the agents. In this paper, we propose a lightweight video semantic segmentation approach---suited to onboard real-time inference---achieving high temporal consistency on aerial data through \textbf{S}emantic \textbf{S}imilarity \textbf{P}ropagation across frames. \textbf{SSP} temporally propagates the predictions of an efficient image segmentation model with global registration alignment to compensate for camera movements. It combines the current estimation and the prior prediction with linear interpolation using weights computed from the features similarities of the two frames. Because data availability is a challenge in this domain, we propose a consistency-aware \textbf{K}nowledge \textbf{D}istillation training procedure for sparsely labeled datasets with few annotations. Using a large image segmentation model as a teacher to train the efficient SSP, we leverage the strong correlations between labeled and unlabeled frames in the same training videos to obtain high-quality supervision on all frames. \textbf{KD-SSP} obtains a significant temporal consistency increase over the base image segmentation model of $12.5\%$ and $6.7\%$ TC on UAVid and RuralScapes respectively, with higher accuracy and comparable inference speed. On these aerial datasets, KD-SSP provides a superior segmentation quality and inference speed trade-off than other video methods proposed for general applications and shows considerably higher consistency. Project page: \href{https://github.com/FraunhoferIVI/SSP}{https://github.com/FraunhoferIVI/SSP}.
\end{abstract}    
\section{Introduction}
%\footnote{$^*$Work done during an internship at Fraunhofer IVI.}
\label{sec:intro}
\begin{figure}[h!]
    \centering
    \includegraphics[width=0.85\linewidth]{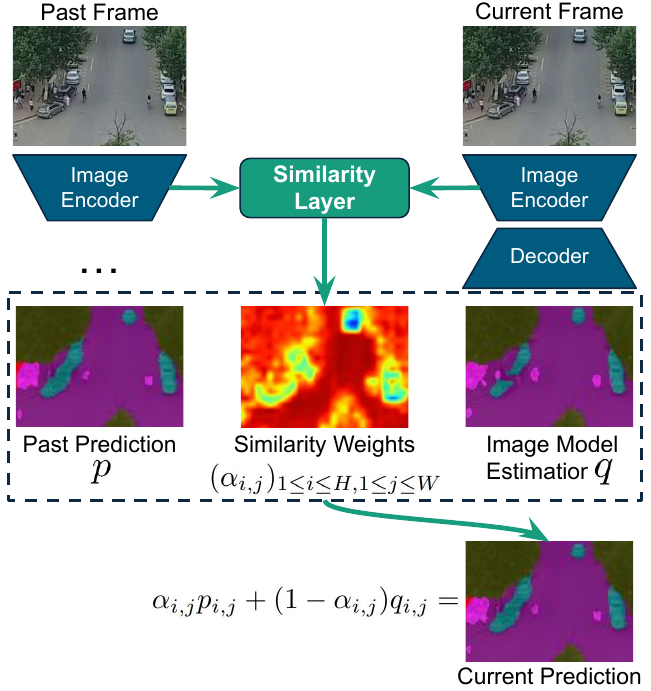} % Replace with your image - could you replace this one with pdf?
    \caption{\textbf{Overview of Semantic Similarity Propagation (SSP)}. On each frame, the estimation of the image segmentation model is linearly combined with the last prediction based on interpolation weights computed from the feature similarities. This results in temporarily consistent robust predictions. The global registration alignment is omitted here.}
    \label{videomodelpresentation}
\end{figure}

Autonomous Unmanned Aerial Vehicles (UAVs) rely on environmental perception for navigation. Semantic segmentation from RGB cameras serves as a crucial component for interpreting their surroundings, with additional use cases in emergency landing \cite{emlHausmann}, 3D semantic reconstruction for scene understanding \cite{app10041275}, or power lines detection \cite{machines10100881powerlines}. It also has applications in other downstream tasks such as surveillance or photogrammetry \cite{islam2023dronesriseexploringcurrent, uavPhotogrammetric9627932}. For trustworthy systems, the predictions must be temporally stable to avoid contradictory information or inconsistent predictions of crucial objects such as obstacles. While an image model can accurately process each frame independently, the temporal consistency of predictions requires a video model leveraging the correlations between consecutive frames. %Additionally, autonomous flight requires on-board, real-time processing of the video stream. 

Aerial footage captured at reasonable altitudes contains vastly different motion content compared to ground-level settings. With objects moving in a plane approximately orthogonal to the camera axis and at a large distance, most of the change between consecutive frames is a product of the camera motion. This results in a highly consistent semantic content requiring a specific temporal modeling method to be fully exploited. As an additional challenge, available data in this domain is scarce, and producing quality annotations of the complex scenes captured at high altitudes is resource-intensive. All publicly available datasets are sparsely labeled with few annotated frames per video: leveraging unlabeled frames in a semi-supervised setting is essential for better generalization.

Autonomous flight requires onboard, real-time processing of the video stream with high accuracy and consistency for reliability. However, most video segmentation approaches compromise on either inference speed or accuracy. Feature enhancement methods reach state-of-the-art accuracy through expensive temporal context modeling not suited to onboard applications \cite{netwarp, miao2021vspw, cffm++}. Alternatively, key frame methods increase inference speed by re-using outputs from a subset of key frames to predict the intermediary non-key frames with fewer computations but inevitably less accuracy \cite{shelhamer2016clockworkconvnetsvideosemantic, zhu2017deepfeatureflowvideo, xu2018dynamicvideosegmentationnetworkdvsnet}. Both approaches are limited in prediction stability, as they do not encourage it in their training or architecture.

% larger segmentation model -> teacher model
Our work aims to achieve reliable predictions---both accurate and stable---by leveraging temporal information across frames without sacrificing efficiency. For a lightweight approach suited to onboard applications, we explicitly \textbf{propagate} predictions to enforce consistency \cite{6630567efficienttempconststream, nilsson2017semanticvideosegmentationgatedgru, TFC}. We linearly combine the output of an image segmentation model on the current frame with the previous frame's prediction at each time step, using per-pixel interpolation weights based on \textbf{semantic similarities} between the frames. Inspired by TFC \cite{TFC}, which uses cosine similarity on the feature maps to compute the weights, we extend their propagation method to entire videos using a convolutional similarity layer capable of learning the optimal balance between the past and current predictions without supervision from the labels. To further enhance consistency, we align the past prediction to the current frame using global registration to compensate for camera motion. This is significantly more efficient than optical flow alignment \cite{6630567efficienttempconststream, nilsson2017semanticvideosegmentationgatedgru} as the registration homography can be estimated from the UAV pose tracked by onboard sensors. %%NEW%%
While global registration assumes a planar ground, we show that our interpolation is robust to imperfect alignments in more complex real-world scenes. We complement our method with \textbf{knowledge distillation} for the semi-supervised training of our efficient model, utilizing a larger teacher model for supervision on all frames. ETC \cite{etc} introduced knowledge distillation for temporal consistency in an efficient image segmentation model, which we adapt to SSP. Instead of jointly training the teacher and student for consistency, we post-process the former's predictions with optical flow and restrain the losses to the prediction level because the teacher model cannot achieve the high consistency of explicitly propagated predictions.

%showed that knowledge distillation can be used to train an efficient image model to higher temporal consistency, and we use the same consistency loss. However, we enforce stability in the architecture of SSP, and instead use knowledge distillation to train on a higher number of samples leading to better generalization and higher accuracy. Because of this, we do not need to jointly fine-tune the teacher model for consistency and instead post-process its predictions with optical flow for consistent supervision.
%While ETC \cite{etc} employed knowledge distillation to train an efficient image model to higher temporal consistency, we instead use it for supervision on a higher number of samples leading to better generalization and higher accuracy.

In summary, we propose a new video semantic segmentation approach suitable for real-time onboard UAV applications without compromising accuracy, consistency, or efficiency. Our contributions are as follows:
\begin{itemize}
    \item We predict linear interpolation weights with a convolutional similarity layer for a consistent and accurate propagation of predictions throughout videos, leveraging the semantic similarities of frames without supervision.
    \item We align previous predictions to the current frame using a global projective transformation, achieving stable combinations in aerial data dominated by camera motion and avoiding the high computational cost of optical flow.
    \item We train with a consistency-aware knowledge distillation, using unlabeled frames to prevent overfitting on the few annotated UAV frames and supervising the temporal propagation of predictions to enhance segmentation accuracy and consistency.
\end{itemize}

% KD-SSP improves the accuracy of the image segmentation model it is based on by 1.4\% mIoU, and reaches 12.51\% higher consistency on the UAVid dataset with a comparable inference speed. Improvements are ? mIoU and ? TC on the RuralScapes dataset. It outperforms feature enhancement methods \cite{netwarp, miao2021vspw} especially in consistency but also in accuracy because of the knowledge distillation, and provides a better quality/efficiency trade-off than key frame methods \cite{zhu2017deepfeatureflowvideo}.

\section{Related work}
\label{sec:relatedwork}
\begin{figure*}[t]
    \centering
    \includegraphics[width=0.95\linewidth]{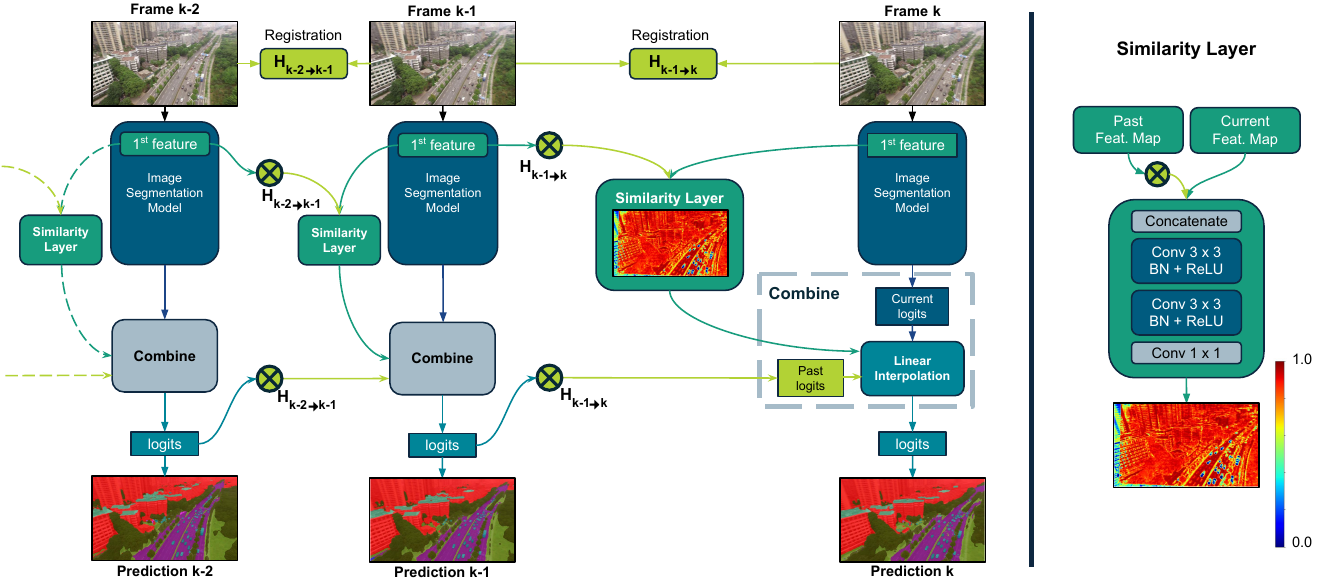} % Replace with your image
    \caption{\textbf{Architecture of SSP, shown during inference over a video.} Predictions (as logits) are propagated through the whole video, with a global projective transformation $H$ alignment to compensate for the camera movements. On each frame, the current estimation of the image semantic segmentation model is combined with the aligned past prediction by linear interpolation. The interpolation weights are computed from the extracted feature maps of the two frames by learnable convolutional layers. The first frame of each video is processed by the image model only. $\otimes$ represents the homography warping operation.}
    \label{videomodelpresentation}
\end{figure*}

% and results in -> resulting in
\paragraph{Image semantic segmentation.} Since the introduction of FCN \cite{fcn}, encoder-decoder architectures with pre-trained backbones have been the standard for semantic segmentation. CNN models have evolved through larger spatial context modeling \cite{deeplab, deeplab3, deeplab3+, zhao2017pyramidsceneparsingnetworkpspppm}, multi-scale representations \cite{upernet}, and relational context aggregation based on pixel correlations rather than spatial proximity \cite{OCR}. Transformers further enhance this with self-attention \cite{10.5555/3295222.3295349transformer, vit}, and hierarchical transformers suited to dense predictions \cite{swin, xie2021segformer, ryali2023hiera} now define state-of-the-art performance. Using image segmentation models on videos ignores the temporal correlations between frames, resulting in low consistency across predictions, especially for sparsely labeled datasets as few training samples lead to overfitting and a lack of regularization.

%can be extended effectively to video semantic segmentation by processing each frame independently and are capable of high accuracy. However, they ignore the temporal correlations in videos and, as a result, have low consistency across predictions . This is accentuated for sparsely labeled datasets with few training samples on which the model can overfit, leading to a lack of regularization and stability. We efficiently and significantly increase the consistency of predictions through the correlations of consecutive frames and regularization in our training.

\paragraph{Video semantic segmentation.} Real-time applications require online inference, which excludes methods leveraging bi-directional temporal interactions \cite{8099707budjetaware, 10484220bofp, tempstable}. Most video methods are based on an encoder-decoder image segmentation model, introducing temporal interactions at the feature or prediction level. The two main approaches are accurate feature enhancement methods and efficient key frame methods. The former reaches higher accuracy by incorporating temporal context from adjacent frames into the static representations of the image encoder: STFCN \cite{fayyaz2016stfcnspatiotemporalfcnsemantic} propagates features with a recurrent module, while NetWarp \cite{netwarp} uses optical flow and linear interpolation. Since TCB \cite{miao2021vspw}, most methods use attention-based mixing modules \cite{tmanet, 10.1145/3474085.3475409stt, dstfp, ssltm, cffm++}. While these methods can scale beyond image models, they do not fit the computational and memory constraints of real-time on-edge inference. Key frame methods instead leverage temporal correlations to avoid redundant computations and achieve faster inference \cite{shelhamer2016clockworkconvnetsvideosemantic, zhu2017deepfeatureflowvideo}. Deep Feature Flow (DFF) \cite{zhu2017deepfeatureflowvideo} re-uses the feature maps of the last key frame aligned with optical flow. Variants include adaptive key frame selection \cite{xu2018dynamicvideosegmentationnetworkdvsnet, li2018lowlatencyvideosemanticsegmentation, articledhnaks, Cai2023AttentionBQqnet}, methods for re-introducing non-key frame information \cite{li2018lowlatencyvideosemanticsegmentation, articledhnaks, jain2019accelcorrectivefusionnetwork}, and optical flow alternatives with intrinsic learnable layers \cite{li2018lowlatencyvideosemanticsegmentation, articledhnaks}. Because these methods have inconsistent latency and unreliable non-key frame predictions, we focus on efficient image models and minimize the computational overhead of the temporal interactions used for consistency gains without compromising accuracy.

\paragraph{Temporal consistency of predictions.} The stability of predictions can be explicitly enforced by combining the current frame’s estimation from the base image segmentation model with those from previous frames. Miksik \etal \cite{6630567efficienttempconststream} proposes a spatially local combination with the previous prediction after optical flow alignment. STGRU \cite{nilsson2017semanticvideosegmentationgatedgru} further models consistency across frame sequences by propagating predictions with a gated recurrent unit and optical flow. However, the consistency gains of these methods are accompanied by a reduction in efficiency, mainly because of the slow optical flow computation. TFC \cite{TFC} improves stability without compromising efficiency by using a simple linear interpolation between the current and past predictions without alignment, weighted by a similarity map computed with the cosine similarity of the low-level feature maps. As TFC is also a key frame method, consistency is only enforced between key frames, and predictions are not propagated through entire videos.

Contrastingly, ETC \cite{etc} processes each frame independently during inference with the image model and introduces temporal consistency during the training only through multiple consistency losses at both prediction and feature levels. Knowledge distillation from a teacher model is used to train efficient compact models. The resulting student model has improved temporal consistency through the regularization of representations over similar frames. As we enforce stability by propagating predictions, we use knowledge distillation to train on unlabeled frames instead.

%Knowledge distillation from a teacher model further enhances efficiency, allowing the student image model to achieve greater temporal consistency by regularizing feature representations across similar frames.

%-------------------------------------------------------------------------
\paragraph{Semi-supervised video semantic segmentation.} Most video semantic segmentation datasets are sparsely labeled, as annotating every frame in a video is prohibitively resource-intensive. This is the case for all UAV datasets, which are additionally of smaller sizes. Semi-supervised methods tackle this challenge by leveraging unlabeled frames during training. Previously seen feature enhancement models train the image encoder on unlabeled adjacent frames to improve generalization and reduce overfitting on the few labeled frames \cite{ssltm, Zhuang_2022_CVPR}. Orthogonally, pseudo-label methods generate supervision signals on unlabeled frames either in an offline manner \cite{pseudolabeloffline1, pseudolabeloffline2, pseudolabeloffline3} or online through self-regularization \cite{zou2021pseudosegdesigningpseudolabels, NEURIPS2020_06964dcefixmatch} or the joint optimization of a teacher network with the same architecture \cite{crosspseudo}. Given that our efficient model focuses on the trade-off between prediction quality and inference speed instead of state-of-the-art accuracy, we propose a knowledge distillation training relying on a larger teacher model without any efficiency constraints. We leverage the temporal correlations between labeled and unlabeled frames in the training videos to provide high-quality supervision through inter-video overfitting. Feature enhancement and pseudo-label methods could then be used to train the teacher model for higher-quality supervision, although we do not experiment with them in this work.

% Because ~ do we keep it here? - other version
% We propose knowledge distillation training relying on a larger teacher model without any efficiency constraints to prioritize the trade-off between prediction quality and inference speed over state-of-the-art accuracy. We leverage the temporal correlations between labeled and unlabeled frames in the training videos to provide high-quality supervision through inter-video overfitting. Feature enhancement and pseudo-label methods could be used to train the teacher model for higher-quality supervision, although we have not experimented with them in this work.

% We remove above if we need more space otherwise keep it.
\section{Methodology}
\begin{figure*}[t]
    \centering
    \includegraphics[width=1\linewidth]{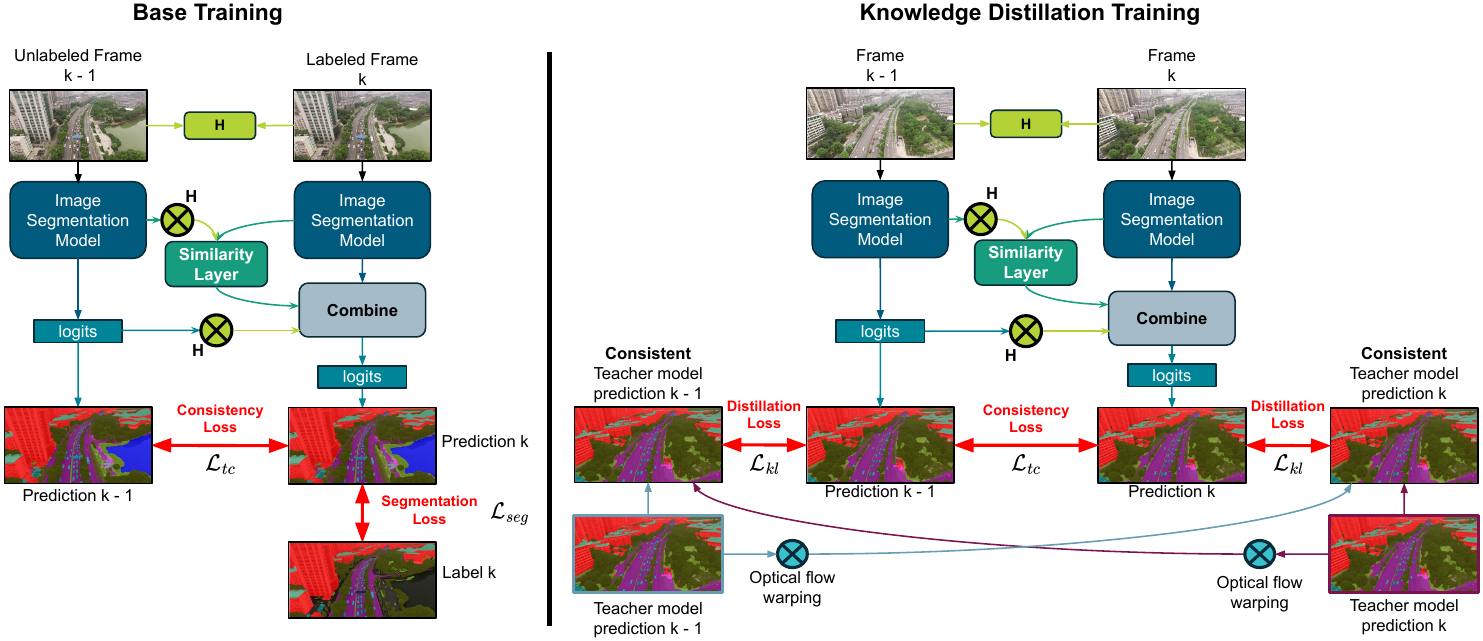} % Replace with your image
    \caption{\textbf{SSP training process.} SSP is trained on two consecutive frames. The prediction of the past frame is propagated to the current frame on which the similarity layer is trained. The consistency loss is computed on the two predictions. Only the current frame needs to be labeled for the base training. With knowledge distillation, we train on all frames and add supervision on the past frame. An optical flow-based post-processing is used to make the teacher model's predictions consistent, as is expected of the student model's.}
    \label{videomodeltraining}
\end{figure*}

The architecture of SSP is illustrated in \cref{videomodelpresentation}. We efficiently augment any image semantic segmentation model for stable inference across complete videos through a simple propagation of predictions with global registration alignment. Predictions on each frame are the linear combination of the image model estimation and the prior knowledge of the past frame's aligned prediction. The result is a more robust estimate for the current frame, more consistent with the past prediction. We propose a temporally consistent knowledge distillation procedure to train this efficient architecture on sparsely labeled datasets.

%Our video model is based on an image segmentation model. After the first frame, predictions are propagated sequentially across the whole video. Global registration is used to align them to the next frame. The prediction of the each frame is a combination of the current estimation from the image model and the last prediction. We propose a knowledge distillation-based training method suited for efficient models and sparsely labeled datasets, with temporally consistent supervision based on the optical flow.

\subsection{Temporal propagation of predictions}
Predictions are propagated as logits. The image model output on the current frame is combined with the last frame prediction only as it is sufficient to enforce stability.
%Predictions are sequentially propagated across the video. On a given frame, the estimation of the base image model is linearly combined to the prediction made on the immediately preceding frame, both as continuous logits. The past prediction is aligned to the current frame with a projective transformation computed from global registration. 
The combination is a simple linear interpolation with weights representing the degree of semantic consistency between the two frames at each pixel's location. 
%These weights are computed by comparing the low-level feature maps of both frames extracted by the base image model. This is a measure of semantic consistency: as the predictions are aligned with global registration compensating only for the camera movements, combined pixels are unlikely to match exactly, yet only the consistency of their semantic content matters for our goals of accuracy and temporal consistency.

\paragraph{Semantic similarity-based linear interpolation.} 
With $p$ the last prediction aligned to the current frame, and $q$ the current output of the image model, the prediction of SSP is:
\begin{equation}
    \hat{q}_{i,j} = \alpha_{i,j} p_{i,j} + (1-\alpha_{i,j}) q_{i,j}
    \label{linearinterpolation}
\end{equation}
where $(\alpha_{i,j})_{1\leq i\leq H,1\leq j\leq W}$ are the weights representing the semantic consistency on each pixel $(i, j)$. These weights are computed by the similarity layer between the low-level feature maps of the past and current frame, illustrated in the right part of \cref{videomodelpresentation}. We use convolutional layers on their concatenation for a local pattern-based detection of similarity between the semantic content of the two frames instead of the per-pixel cosine similarity of TFC \cite{TFC}. Using more expressive learnable layers instead of a direct similarity measure facilitates the optimization of the predicted weights and minimizes the influence on the image encoder, which can focus on producing features suited to the segmentation task.

%\noindent This is a fusion of two estimations for the segmentation of the current frame: the current prediction from the image segmentation model and the prior knowledge from the past prediction aligned to the current frame. The result is a more robust estimate for the current frame, consistent with the past prediction. 

%The similarity-based interpolation fuses the past and current predictions to obtain the improved current prediction. This fusion is a linear interpolation with per-pixel weights obtained from the similarity layer:
%$$\hat{y}_{i,j} = \alpha_{i,j} x_{i,j} + (1-\alpha_{i,j}) y_{i,j}$$
%where $x$ is the last prediction and $y$ is the current one from the image model. 
%The similarity layer predicts the $(\alpha_{i,j})_{i,j}$ by comparing the first low-level feature map of both frames, with the past one aligned with the same homography as the prediction.  

%\subsubsection{Optional Prediction-Level Mixing}
%Prediction mixing is a spatially local mutual enhancement of both frames based on their correlations. This module can handle misalignments and enlarges the expressivity of the model as it can represent relations involving more than the two corresponding pixels. This step updates the past and current predictions but does not fuse them. 
%This module is optional: while it can improve results, it is computationally expensive relatively to its benefits.
%The similarity layer can include a feature mixing step similar to prediction mixing to handle misalignments and update the features to prepare them for the similarity computation.

\paragraph{Global registration alignment.}
For the linear interpolation combination, the past prediction and feature map are aligned to the current frame with global registration to compensate for the camera movement only. We consider a homography $H$ representing a projective transformation:

\begin{equation}
    \begin{bmatrix}
    x' \\
    y' \\
    1
    \end{bmatrix}
    =
    H
    \begin{bmatrix}
    x \\
    y \\
    1
    \end{bmatrix}
    =
    \begin{bmatrix}
    h_{11} & h_{12} & h_{13} \\
    h_{21} & h_{22} & h_{23} \\
    h_{31} & h_{32} & 1
    \end{bmatrix}
    \begin{bmatrix}
    x \\
    y \\
    1
    \end{bmatrix}
    \label{alignment}
\end{equation}
where $x$ and $y$ are the coordinates in the original frame, and $x'$ and $y'$ are the coordinates in the aligned image. %%%NEW%%% 
This transformation assumes a planar ground surface, which will result in alignment errors in more complex scenes. This does not result in a loss of accuracy as the similarity layer already expects misaligned pixels due to moving objects. Aligning the combined frames, even roughly, enables more pixels to corresponds and leads to higher consistency.

In practice, $H$ is computed from the displacement (rotation and translation) of the UAV between the two frames, assuming that the ground is a flat plane and that the altitude is known \cite{Hartley_Zisserman_2004multipleview, 5548048homography}. As this is unavailable in the considered datasets, we use a feature-based estimator matching keypoints between the frames \cite{tyszkiewicz2020disklearninglocalfeatures, lindenberger2023lightglue} for our experiments. This estimator and the differentiable warping operation are implemented by the Kornia library \cite{riba2019korniaopensourcedifferentiable}.
% based on henri's feedback we can make sound obvious?
% In practice, $H$ is computed based on the estimated UAV pose and altitude, assuming a horizontal ground plane for simplicity.

\subsection{Video training}
We first present a basic training procedure for SSP without knowledge distillation illustrated in \cref{videomodeltraining}, relying only on labeled samples. Training is done on two consecutive frames, propagating the prediction of the past frame $k-1$ to the current one $k$ on which the combination layers are trained. Only the current frame is assumed to be labeled, and the segmentation loss is computed on its prediction. To train for the temporal consistency of predictions in an unsupervised manner, as the past frame is not labeled, we use the same consistency loss as ETC \cite{etc}:
\begin{equation}
    \mathcal{L}_{tc} = \frac{1}{HW}\sum_{i,j} \textbf{O}^{k-1\to k}_{i,j}\left\|y_{i,j} - \hat{x}_{i,j} \right\|_2^2
    \label{etcloss}
\end{equation}
where $y$ is the current prediction, $\hat{x}$ the past prediction warped to the current frame with optical flow, and $\textbf{O}^{k-1\to k}$ an occlusion mask to reduce the influence of flow estimation errors and handle pixels which cannot be matched by optical flow: $\textbf{O}^{k-1\to k}_{i,j} = exp(-\left\|\textbf{I}^k_{i,j} - \hat{\textbf{I}}^{k-1}_{i,j}\right\|_1)$
%\begin{equation}
%    \textbf{O}^{k-1\to k}_{i,j} = exp(-\left\|\textbf{I}^k_{i,j} - \hat{\textbf{I}}^{k-1}_{i,j}\right\|_1)
%    \label{occlusionmask}
%\end{equation}
, with $\hat{\textbf{I}}^{k-1}$ the past frame warped with optical flow to the current frame $\textbf{I}^k$.

%integrate it in the training objective. The consistency loss is unsupervised as the past frame's label is not available. It is computed between the two predictions based on consistency information from the input frames. Using only two frames during training relieves the memory requirements compared to training on longer sequences, and is efficient as all necessary information for temporal consistency is in the immediate past frame.

As the base image model is not frozen, it is trained on both frames from only the current frame's label. Similarly to ETC \cite{etc}, this increases regularization through the consistency loss that ensures similar predictions on the two close frames. The total loss is $\mathcal{L}(P_{q}, P_{p}, A_{q}) = \mathcal{L}_{seg}(P_{q}, A_{q}) + \lambda \mathcal{L}_{tc}(P_{q}, P_{p})$, where $P_{q}$ and $P_{p}$ are the current and past predictions respectively, and $A_{q}$ the ground truth for the current frame. $\mathcal{L}_{seg}$ is the cross-entropy loss, and $\lambda$ is the consistency loss weight.

%The segmentation loss is the cross-entropy. The base segmentation model is trainable and not frozen, and the backpropagation of both losses goes through its activations on both frames equally. This means that the model is trained on the previous frame despite the absence of a label. In our experiments, the base image model is expected to have been trained beforehand with the image segmentation task on the dataset, as it is used as the image baseline for the task.
 
%The total loss function of the video model is:
%\begin{equation}
%    \mathcal{L}(P_{q}, P_{p}, A_{q}) = \mathcal{L}_{seg}(P_{q}, A_{q}) + \lambda \mathcal{L}_{tc}(P_{q}, P_{p})
%    \label{totallossbase}
%\end{equation}
%where $P_{q}$ and $P_{p}$ are the current and predictions respectively, $A_{q}$ the ground truth for the current frame. $\mathcal{L}_{seg}$ is the segmentation loss, which is the cross-entropy loss for the base training. $\mathcal{L}_{tc}$ is the consistency loss with $\lambda$ its weight.

% segmentation deleted since it is obvious that it is a segmentation model
% this large model -> teacher model
\subsection{Temporally consistent knowledge distillation}
For semi-supervised training on sparsely labeled datasets, we propose to train on all frames through the supervision of a teacher model. This is a larger and more accurate image model trained on the annotated samples, which has no real-time inference efficiency constraints. %We only consider a larger model with a higher input resolution for simplicity, as this is sufficient to reach considerably higher accuracy. 
%overfitting to annotated samples
We leverage the high similarity between frames of the same video to overfit this larger model to the training videos and produce high-quality soft annotations on all frames, unlabeled or not. 

Our temporally consistent knowledge distillation training for SSP is illustrated in \cref{videomodeltraining}. We use the densely available teacher model predictions to supervise both frames with the distillation loss. However, as we also train for consistency, these predictions must be post-processed to be consistent. With $T_{q}$ and $T_{p}$ the teacher model's prediction on the current and past frames respectively, we use the optical flow between the two frames in both directions to obtain the consistent predictions used for supervision $T^c_{q}$ and $T^c_{p}$:
\begin{align}
    T^c_{p} &= (T_{p} + (1-M_{occ})(W_{q\to p} * T_{q}))\frac{1}{2 - M_{occ}}\\
    T^c_{q} &= (T_{q} + (1-M_{occ})(W_{p\to q} * T_{p}))\frac{1}{2 - M_{occ}}
    \label{consistentteacher}
\end{align}
where $W_{q\to p}$ and $W_{p\to q}$ are the optical flows between the frames in each direction, and $*$ is the optical flow warping operation. $M_{occ}$ is a binary occlusion mask computed from a consistency check between the forward and backward flows \cite{tempstable, Ruder_2016occlusionoptflow}: $M_{occ} = \left\| W_{q\to p} + W_{p\to q} \right\|_2^2 > 0.01(\left\| W_{q\to p} \right\|_2^2 + \left\| W_{p\to q} \right\|_2^2) + 0.5$.
%\begin{equation}
%    \small
%    M_{occ} = \left\| W_{q\to p} + W_{p\to q} \right\|_2^2 > 0.01(\left\| W_{q\to p} \right\|_2^2 + \left\| W_{p\to q} \right\|_2^2) + 0.5
%    \label{flowoccmask}
%\end{equation}

% we follow the knowledge distillation training introduced in \cite{hinton2015distillingknowledgeneuralnetwork} with the Kullback–Leibler divergence as a loss function and a temperature parameter $\tau=2$.

For the segmentation loss, we use the knowledge distillation approach from \cite{hinton2015distillingknowledgeneuralnetwork}, applying Kullback–Leibler divergence with a temperature parameter $\tau=2$. The true labels are discarded, the teacher model is frozen, and we only supervise at the logits level: $\mathcal{L}_{kl}(P, T) = \text{softmax}(T/\tau)\log\left(\frac{\text{softmax}(T/\tau)}{\text{softmax}(P/\tau)}\right) \times \tau^2$, 
%\begin{equation}
%    \mathcal{L}_{kl}(P, T) = \text{softmax}(T/\tau)\log\left(\frac{\text{softmax}(T/\tau)}{\text{softmax}(P/\tau)}\right) \times \tau^2
%    \label{kdloss}
%\end{equation}
where $P$ and $T$ are the student and teacher logits, respectively. The distillation loss cannot be used at the feature level, as the consistency of the teacher model is not guaranteed. The loss function of the video knowledge distillation training is $\mathcal{L}(P_{q}, P_{p}, T^c_{q}, T^c_{p}) = \mathcal{L}_{kl}(P_{q}, T^c_{q}) + \mathcal{L}_{kl}(P_{p}, T^c_{p}) + \lambda_{kd} \mathcal{L}_{tc}(P_{q}, P_{p})$.

\section{Experiments}
% I think it would be nice if we can add few lines here what wer are doing here?
% Base image model: 310.6 vs. SSP (ours): 322.8 vs. NetWarp: 739.9 vs. TCB: 1350.3 GFLOPs
\begin{table*}[t!]
\small
\centering
\begin{tabular}{|c|c|c|c|cc||cc||cc|}
\hline
\multirow{2}{*}{} & \multirow{2}{*}{\textbf{Method}} & \multirow{2}{*}{\textbf{Params}} & \multirow{2}{*}{\textbf{GFLOPs}} & \multicolumn{2}{c||}{\textbf{FPS}} & \multicolumn{2}{c||}{\textbf{UAVid}} & \multicolumn{2}{c|}{\textbf{RuralScapes}} \\
                  &                    &                 & & \textbf{A100}  & \textbf{Orin} & \textbf{mIoU $\boldsymbol{\uparrow}$}  & \textbf{TC $\boldsymbol{\uparrow}$}       & \textbf{mIoU $\boldsymbol{\uparrow}$} & \textbf{TC $\boldsymbol{\uparrow}$}  \\ 
\hline
 & \cellcolor{lightgray}\textbf{Teacher Model} & \cellcolor{lightgray}\textbf{101.01M} & \cellcolor{lightgray}- & \cellcolor{lightgray}- & \cellcolor{lightgray}- & \cellcolor{lightgray}\textbf{81.92} & \cellcolor{lightgray}\textbf{84.09} & \cellcolor{lightgray}\textbf{66.65} & \cellcolor{lightgray}\textbf{89.43}\\
                                                               & SegFormer - b2 & \textbf{27.36M} & 204.1 & 77 & - & 77.81    & 83.76  & 62.75 & 86.89\\
 \centering\arraybackslash \textbf{Image}   & SegFormer - b3 & 47.23M & 256.7 & 48 & - & 78.02    & 82.59  & 63.53 & 86.65\\
 \centering\arraybackslash \textbf{Models}  & ConvNeXt-S + UPerNet & 81.77M & 922.0 & 96 & - & 78.35    & 83.13  & 63.29 & 86.70\\
                                                               & \cellcolor{gray!10}Base Image Model &  \cellcolor{gray!10} & \cellcolor{gray!10} &  \cellcolor{gray!10} &  \cellcolor{gray!10} & \cellcolor{gray!10}79.23 & \cellcolor{gray!10}79.02 & \cellcolor{gray!10}63.51 & \cellcolor{gray!10}87.34\\
                                                               & \cellcolor{gray!10}KD Base Image Model (Ours) & \cellcolor{gray!10}\multirow{-2}{*}{43.17M} & \cellcolor{gray!10}\multirow{-2}{*}{310.6} & \cellcolor{gray!10}\multirow{-2}{*}{\textbf{104}} & \cellcolor{gray!10}\multirow{-2}{*}{\textbf{31.4}} & \cellcolor{gray!10}80.38 & \cellcolor{gray!10}87.15 & \cellcolor{gray!10}64.46 & \cellcolor{gray!10}90.37\\
\hline  
                                                            & DFF \cite{zhu2017deepfeatureflowvideo} & 48.43M & \textbf{137.2} & 23$^*$ & - & 77.20    & 83.28  & 62.66 & 88.75\\
                                                            & NetWarp \cite{netwarp} & 48.44M & 739.9 & 15$^*$ & - & 79.31 & 82.19  & 63.99 & 88.48\\
 \centering\arraybackslash \textbf{Video}                   & $\text{TCB}_{ppm}$ \cite{miao2021vspw} & 64.56M & 1350.3 & 19 & - & 79.61 & 81.35  & 63.83 & 87.73\\
 \centering\arraybackslash \textbf{Models}                  & $\text{TCB}_{ocr}$ \cite{miao2021vspw} & 63.49M & 1379.4 & 18 & - & 79.67 & 82.22 & 63.56 & 88.39\\
 
                                                            & \cellcolor{gray!10}SSP (Ours) & \cellcolor{gray!10}  & \cellcolor{gray!10} & \cellcolor{gray!10} & \cellcolor{gray!10} & \cellcolor{gray!10}79.75 & \cellcolor{gray!10}\textbf{92.10}  & \cellcolor{gray!10}64.00 & \cellcolor{gray!10}\textbf{94.06}\\ %79.87/79.53/79.85, 92.01/92.27/92.03, ..., 64.03/64.34/63.63, 94.01/94.07/94.11
                                                            %& Video-mixing & 43.47M & 73 & 80.04/79.31 & 91.54/92.03  & 64.40/63.86 & \textbf{94.26}/93.99\\
                                                            & \cellcolor{gray!10}KD-SSP (Ours) & \cellcolor{gray!10}\multirow{-2}{*}{43.38M} & \cellcolor{gray!10}\multirow{-2}{*}{322.8} & \cellcolor{gray!10}\multirow{-2}{*}{95} & \cellcolor{gray!10}\multirow{-2}{*}{29.3} & \cellcolor{gray!10}\textbf{80.63} & \cellcolor{gray!10}91.53  & \cellcolor{gray!10}\textbf{64.56} & \cellcolor{gray!10}94.00\\%80.54/80.66/80.70, 91.43/91.59/91.58, ..., 64.28/65.04/64.37, 93.96/93.98/94.06
                                                            %& KD-Video-mixing & 43.47M & 73 & \textbf{80.66}/80.87 & 91.55/91.66  & 64.98/64.62 & 93.72/93.78\\
\hline
\end{tabular}
\caption{\textbf{Comparison on the validation sets of aerial datasets UAVid and RuralScapes.} Base Image Model is Hiera-S and UPerNet, Teacher Model is Hiera-B+ and UPerNet. Input resolution is $736\times 1280$ except for the teacher model. While the feature enhancement methods outperform the base image model, SSP has the highest mIoU and TC. KD-SSP has a higher mIoU than the base image model trained with knowledge distillation (KD). Reported metrics on SSP and KD-SSP are the average of 3 runs. $^*$ For the optical flow computation in these models, we used the accurate but relatively slow RAFT \cite{teed2020raftrecurrentallpairsfield}, in the small configuration with 8 iterations. While a faster model would lead to higher inference speed, especially for DFF, it would result in even lower mIoU for DFF and an even more negligible improvement in mIoU and TC over the base image model for NetWarp.}
\label{resultstable}
\end{table*}
\subsection{Datasets}
We conduct our experiments on two UAV video semantic segmentation datasets. UAVid \cite{uavid} is an urban dataset with 27 video sequences of 901 frames, captured at an altitude of around 50 meters and 20 frames per second (FPS) for 45 seconds each. It contains 270 annotated images, one every 100 frames or 5 seconds. Because each labeled sample requires a past frame to train SSP, we discard the first annotated frame of all videos and use only 243 annotated images. This corresponds to 180/63 labeled images and 18000/6300 total frames for training and validation, respectively. We use the annotations from \cite{cai2023vdd}, which include the addition of ``roof" and ``water" classes and the merging of ``moving car" and ``static car" for 8 semantic classes in total, to make this dataset more uniform with others for downstream applications. This re-mapping is only available for training and validation splits. Background is ignored during training. %(because too diverse + close to other classes) 
The RuralScapes dataset \cite{marcu2020semanticssegpropruralscapes} is of comparable size to UAVid, with a focus on rural scenes. It is more densely annotated with one labeled frame per second for 1127 annotations. We sample this dataset at 10 FPS. As previously, we discard the first annotated frame of each video for training. This corresponds to 791/297 labeled images and 7139/3000 total frames for training and validation, respectively. RuralScapes contains 12 semantic classes different than those of urban scenes.

%There is a significant difference between the distributions of the classes "forest" and "hill" between the training and test splits \ref{ruralscapesclassditribution}, which is an issue as these two classes are easily confused, even for human annotators (because they are not exclusive: most hills have "forests" on them).

\subsection{Evaluation metrics}
Segmentation accuracy is measured with the mean Intersection over Union (mIoU) \cite{fcn}. For stability, we require an unsupervised metric for sparsely labeled datasets and use the Temporal Consistency (TC) \cite{etc}. Given a pair of consecutive frames at time steps $k-1$ and $k$, it is defined as the mIoU between the first prediction $P_{k-1}$ warped by optical flow $W_{k-1\to k}$ to the second frame, and the second prediction $P_{k}$: $TC_{k-1, k} = mIoU(W_{k-1\to k} * P_{k-1}, P_{k})$
%\begin{equation}
%    TC_{k-1, k} = mIoU(W_{k-1\to k} * P_{k-1}, P_{k})
%    \label{tcequation}
%\end{equation}
, where $*$ is the optical flow warping operation. Then, the TC of a video is the mean of the TC of all consecutive pairs of frames.
%$$TC= \frac{1}{N-1}\sum_{t=1}^{N-1}TC_{t, t+1}$$
The inference speed is reported in frames per second (FPS) on an A100 (80G) and a Jetson Orin AGX (32G) GPU over $1000$ samples after warmup, and only includes the forward pass. The global registration estimation is ignored for the speed benchmark of SSP, as it would be derived from the UAV pose in practice. We similarly report GFLOPs for an estimation of the computational cost.
% Here i think pose more accurate term? anyway estimating the H from the "Estimiate" pose sounds cool

% classification logits for each pixel -> pixel-wise classification 
% to judge the benefits of our video method on a precise model ->, allowing us to evaluate our video method's benefits,
\subsection{Experimental details}
\paragraph{Base image model.} Any image semantic segmentation model producing pixel-wise classification logits can be used. We consider an encoder-decoder with Hiera-S \cite{ryali2023hiera} as the backbone, and UPerNet \cite{upernet} with $256$ channels for the segmentation head.  We use the implementation and pre-trained weights from the Segment Anything 2 (SAM2) encoder \cite{ravi2024sam2}. The results of the base image model serve as a baseline in our experiments. This model was chosen to obtain high segmentation accuracy efficiently on both datasets (see the comparison of image models in \cref{resultstable}), allowing us to evaluate our video method's benefits on a precise model, as temporal consistency is intuitively harder to obtain on detailed predictions.

%Student models in knowledge distillation are trained with the same parameters, for $20$ and $25$ epochs on UAVid and RuralScapes respectively.
%\paragraph{Teacher model.} For simplicity, we consider the same architecture in a larger size: Hiera-base+ and UPerNet with $512$ channels. More importantly, this model uses a larger input resolution of $1440\times 2560$. 

\paragraph{SSP.} SSP is based on the base image model. We use the trained image models as starting points for our experiments to ensure easier and faster convergence, but this choice is challenged in the ablation study. SSP has 43.47M parameters, only $0.7\%$ more than the base image model. For the similarity layer, we use the first extracted feature map of stride $4$. Each convolutional layer divides the number of channels by 2. We set $\lambda=0.5$ for the base training, and $\lambda_{kd}=135000$.

\paragraph{Training parameters and teacher model.}
Except for the teacher model, the input resolution is set at $736\times 1280$. The base image model is trained for $120$ and $60$ epochs on UAVid and RuralScapes respectively, with a batch size of $4$, a learning rate of $2.5\cdot 10^{-5}$, an AdamW optimizer with cosine scheduler and warmup, and $0.05$ weight decay. For data augmentation, we use horizontal flipping, random scale and rotations, random noise, and random brightness/color changes. SSP is trained for $200$ and $75$ epochs on each dataset respectively, with double the learning rate and random square cropping to the shorter side in order to save memory. Other parameters are the same. For knowledge distillation, we use the same parameters and train for $20$ and $25$ epochs on UAVid and RuralScapes respectively.

The teacher model is trained for $200$ and $75$ epochs respectively, with the same parameters and random square cropping. For simplicity, we consider the same architecture in a larger size: Hiera-B+ and UPerNet with $512$ channels, and with a larger input resolution of $1440\times 2560$.

%The models are trained for $200$ and $75$ epochs on the UAVid and RuralScapes datasets, respectively. All training parameters are the same. Random square cropping to the shorter side is used to save memory.

%For UAVid, knowledge distillation samples are chosen as 1 every 4 frames. This results in 25 samples per 100-frames clip, and the video model is trained on 50 frames in total. This halves the number of saved teacher predictions on disk and avoids similar consecutive frames in the same training mini-batch. For RuralScapes, samples are chosen as 1 every 2 frames. This results in 25 samples per 50-frames clip, and the video model is trained on every frame. We avoid using the same frame as the past frame in one sample and the current frame in another: each frame is exclusive to one training sample. Image models are trained on the same samples as the video models, meaning 1 in every 4 frames for UAVid and 1 in every 2 frames for RuralScapes. This was an arbitrary choice to increase the diversity in mini-batches but might not be optimal.

\subsection{Results}
We compare SSP to the base image model serving as a baseline and other video semantic segmentation methods \cite{zhu2017deepfeatureflowvideo, netwarp, miao2021vspw} all using the same base encoder and decoder architecture. %Details on their implementation can be found in the supplementary (\cref{othermethodssupp}). 
Results on both datasets are reported in \cref{resultstable}. SSP with the base training yields $0.5\%$ mIoU and $13\%$ TC increases over the base image model on UAVid. On RuralScapes, results are similar with $+0.5\%$ mIoU and $+6.7\%$ TC. With knowledge distillation, KD-SSP obtains an additional $+0.8\%$ and $+0.56\%$ mIoU on UAVid and RuralScapes, respectively, for slightly lower TC than the base training. %We observed higher variance in training on RuralScapes (standard deviation of $0.3$ vs. $0.1$ on UAVid), likely due to labeling issues, particularly class confusions between "forest" and "hill," resulting in divergent training and validation distributions. More details can be found in the supplementary (\cref{ruralscapessupp}).

% We note that we obtained higher variance in our training runs on RuralScapes ($0.3$ standard deviation compared to $0.1$ on UAVid), most likely due to its labeling issues consisting mainly of confusions and inconsistencies between the two classes ``forest" and ``hill" which lead to different training and validation distributions. More details can be found in the supplementary (\cref{ruralscapessupp}).

We test the influence of knowledge distillation for semi-supervised learning by training the base image model with it. Results are referred to as the KD base image model. This corresponds to training on all frames from the distillation loss with the teacher model predictions. Temporal consistency is considerably improved, especially on UAVid which is more sparsely labeled, as training on a larger number of frames brings regularization for more uniform predictions on similar frames. Segmentation accuracy is also improved but still lower than KD-SSP, which also outperforms all other video models in mIoU and shows high temporal consistency due to the propagation of predictions. 

SSP only has a marginally lower inference speed than the image model, resulting in a largely superior metrics/inference speed trade-off that is increased further for KD-SSP. Other video methods with worse metrics on these aerial datasets have considerably lower inference speeds due to expensive feature mixing for TCB and slow optical flow computation for DFF and NetWarp. The latter is because an accurate method of processing the frames at high resolution is needed for the small moving objects.

\begin{figure*}[t]
    \centering
    \includegraphics[width=0.95\linewidth]{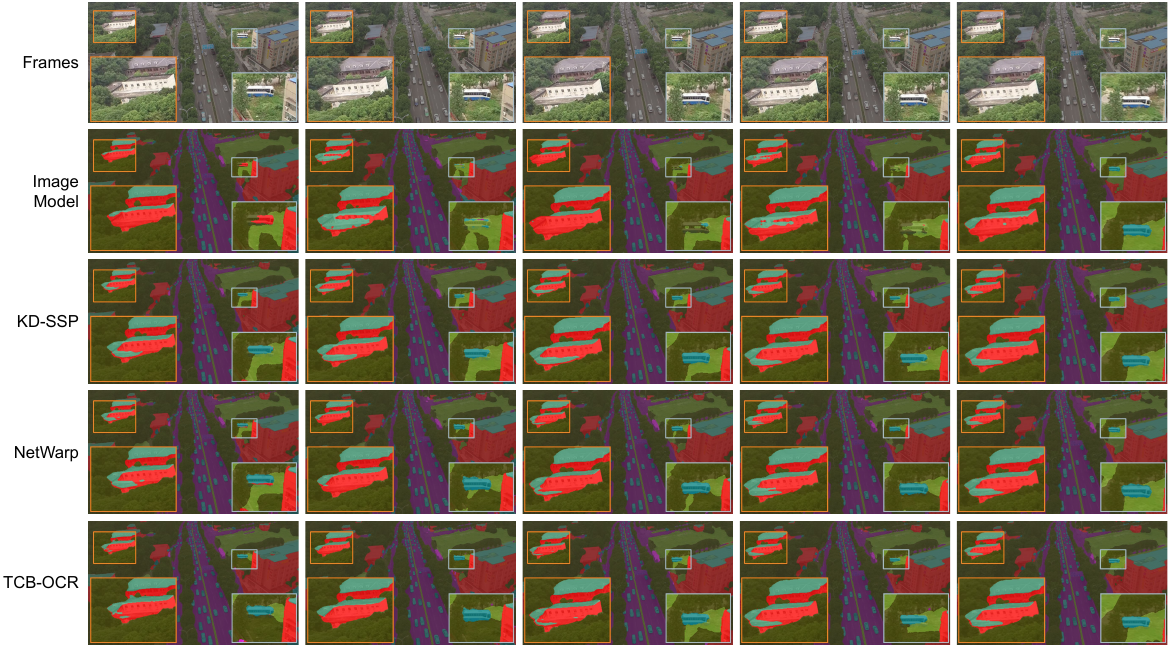} % Replace with your image
    \caption{\textbf{Qualitative results.} Outputs of SSP compared to the base image model and other video methods \cite{netwarp, miao2021vspw} on one video of the UAVid validation set, $736\times 1280$ resolution. Frames are selected 5 frames apart. A comparison of results on entire videos showcasing the consistency gains are included in the supplementary material.}
    \label{qualitativeresults}
\end{figure*}

\subsection{Ablation study}
\begin{table}
    %\small
    \centering
    \begin{tabular}{@{}c|c|cc@{}}
    \hline
    \multicolumn{2}{@{}c}{\textbf{Model}} & \textbf{mIoU} & \textbf{TC} \\
    \hline
    \multicolumn{2}{@{}c}{Base image model} & 79.23 & 79.02 \\
    \hline
    \multirow{4}{*}{SSP} & $\lambda=0.2$ & 79.50 & 90.72 \\ 
                                & $\lambda=0.35$ & \textbf{79.98} & 91.58 \\ 
                                & \cellcolor{gray!10} $\lambda=0.5$ & \cellcolor{gray!10} 79.87 & \cellcolor{gray!10} 92.01 \\
                                & $\lambda=0.65$ & 79.47 & \textbf{92.18} \\ 
    \hline
    \end{tabular}
    \caption{\textbf{Results with different weights of the consistency loss for our method.} Other experiments use $\lambda=0.5$. UAVid dataset, $736\times 1280$ resolution.}
    \label{consistencylossresults}
\end{table}

\begin{table}
    \centering
    %\small
    \begin{tabular}{@{}c|c|cc@{}}
    \hline
    \multicolumn{2}{@{}c|}{\textbf{Model}} & \textbf{mIoU} & \textbf{TC} \\
    \hline
    \multicolumn{2}{@{}c|}{Base image model} & 79.23 & 79.02 \\
    \hline
    \multirow{7}{*}{SSP} & \cellcolor{gray!10} Base parameters  & \cellcolor{gray!10} 79.87 & \cellcolor{gray!10} \textbf{92.01} \\
                                 & Cos. Sim. Interpolation (TFC) & 79.20 & 91.67 \\
                                & No registration alignment & 79.84 & 89.65 \\
                                & No interpolation & 79.99 & 88.20 \\ 
                                & No consistency loss & \textbf{80.01} & 86.38 \\
                                %& No interpolation/consistency loss & 80.31 & 82.55 \\
                                & 1-step training & 79.92 & 91.97 \\
    \hline
    \end{tabular}
    \caption{\textbf{Ablation study of SSP.} \textit{Base parameters} corresponds to the described SSP configuration used for \cref{resultstable}. Results are obtained without knowledge distillation. UAVid dataset, $736\times 1280$ resolution.}
    \label{ablationstudy}
\end{table}

% for quicker experiments -> to expedite the experimental process
We conduct an ablation study of our method on UAVid with the same parameters, omitting knowledge distillation to expedite the experimental process.

\paragraph{Consistency loss weight.} The parameter $\lambda$ balances segmentation and consistency losses. Results with different values of $\lambda$ are in \cref{consistencylossresults}. Values between $0.35$ and $0.5$ appear optimal, as lower or higher $\lambda$ result in worse metrics. This parameter does not control a simple trade-off between mIoU and TC as lower values lead to worse accuracy; however, it can be tuned to change the balance between the two as seen with $\lambda=0.35$ and $\lambda=0.5$.

\paragraph{Similarity layer.} Using cosine similarity as TFC \cite{TFC} instead of our proposed convolutions in the similarity layer, we obtain considerably inferior results and confirm that our choice is better suited to ensure long-term consistency through complete videos without loss of accuracy. Convolutional layers can learn to detect movements and change through local interactions, while the fixed per-pixel similarity measure is limited by the extracted feature maps.

%These layers are superior in learning the optimal balance between the frames to enforce consistency , and in detecting movements and change with local interactions instead of the per-pixel similarity computation.
%he second has spatial filters capable of extracting relevant patterns through linear operations

\paragraph{Global registration alignment.} Global registration is used to compensate for camera movements. Without it, SSP reaches the same mIoU and still significantly outperforms other methods in consistency. Less pixels are matching between the two frames without alignment, which explains the loss of TC. This experiment proves that SSP is robust to errors in the homography estimation as the similarity weights can adapt to misaligned frames and avoid loss of accuracy. SSP is still relevant when homography estimation from the UAV pose is unavailable and its computation from the frames is too slow, as it still outperforms all other methods in TC and mIoU.

\paragraph{Accuracy and consistency trade-off.} We test the influence on accuracy of the two parts of our method enforcing consistency: the linear interpolation for the propagation of predictions and the consistency loss. Without interpolation, corresponding to training the base image model on pairs of frames with consistency loss, our method reaches comparable mIoU with an expected drop in TC. This proves that linear interpolation does not trade-off accuracy for consistency, and that training an image model with the consistency loss already presents high consistency gains as proposed by ETC \cite{etc}. Removing the consistency loss leads to the same conclusion, with even lower TC, showing that this loss is essential to learning temporally consistent predictions. 

\paragraph{Base image model training.} In all experiments, we used the trained base image model as the starting point for SSP, following a 2-step training: first, train the image model, then train SSP as proposed. This ensures easier and faster convergence without any compromise in the case of our experiments. However, in practice, training SSP in one step is more time-efficient, and we prove that this leads to the same metrics without additional training epochs. %This shows that the mIoU increase over the base image model is not simply a result of more training.
\section{Conclusion}
This paper presents SSP, a new method for real-time aerial video semantic segmentation on-edge with high temporal consistency without loss of accuracy. SSP is suited to the motion content of drone-captured videos and to the computational restrictions of autonomous flight. The propagation of predictions with linear interpolation adds minimal computations on top of the base image segmentation model, for considerable gains in segmentation accuracy and especially temporal consistency. Global registration alignment is particularly suited to autonomous flight where it can be estimated from the UAV pose, and where compensating the camera movements leads to increased stability. If unavailable, SSP still outperforms existing image and video methods in accuracy, consistency, and metrics/inference speed trade-off. Alongside this model, we propose a temporally consistent knowledge distillation method for training SSP in a semi-supervised setting and confirm its relevance for efficient models by obtaining notable improvements in segmentation accuracy.

% maybe saying significant improvement in segmentation accuracy sound a bit overly assertive.
{
    \small
    \bibliographystyle{ieeenat_fullname}
    \bibliography{main}
}


\end{document}